\documentclass{ieeeaccess}
\usepackage{cite}
\usepackage{amsmath,amssymb,amsfonts}
\usepackage{algorithmic}
\usepackage{graphicx}
\usepackage{textcomp}
\usepackage{multirow}
\usepackage{booktabs}
\usepackage{multirow}
\usepackage{amsfonts}
\usepackage{tablefootnote}
\usepackage{bm}
\makeatletter
\AtBeginDocument{\DeclareMathVersion{bold}
\SetSymbolFont{operators}{bold}{T1}{times}{b}{n}
\SetSymbolFont{NewLetters}{bold}{T1}{times}{b}{it}
\SetMathAlphabet{\mathrm}{bold}{T1}{times}{b}{n}
\SetMathAlphabet{\mathit}{bold}{T1}{times}{b}{it}
\SetMathAlphabet{\mathbf}{bold}{T1}{times}{b}{n}
\SetMathAlphabet{\mathtt}{bold}{OT1}{pcr}{b}{n}
\SetSymbolFont{symbols}{bold}{OMS}{cmsy}{b}{n}
\renewcommand\boldmath{\@nomath\boldmath\mathversion{bold}}}
\makeatother

\def\BibTeX{{\rm B\kern-.05em{\sc i\kern-.025em b}\kern-.08em
    T\kern-.1667em\lower.7ex\hbox{E}\kern-.125emX}}

\begin{document}
\history{Date of publication xxxx 00, 0000, date of current version xxxx 00, 0000.}
\doi{10.1109/ACCESS.2023.1120000}

\title{MentalQA: An Annotated Arabic Corpus for Questions and Answers of Mental Healthcare }
\author{\uppercase{Hassan Alhuzali}\authorrefmark{1}, 
\uppercase{Ashwag Alasmari}\authorrefmark{2}, and Hamad Alsaleh\authorrefmark{3}}

\address[1]{College of Computing, Department of Computer Science and Artificial Intelligence, Umm Al-Qura University, Makkah, Saudi Arabia (e-mail: hrhuzali@uqu.edu.sa).}
\address[2]{Department of Computer Science, King Khalid University, Abha, Saudi Arabia (e-mail: aasmry@kku.edu.sa).}
\address[3]{College of Computer and Information Sciences, Department of Information System, King Saud University, Riyadh, Saudi Arabia (e-mail: haalsaleh@ksu.edu.sa).}

\markboth
{Author \headeretal: Preparation of Papers for IEEE TRANSACTIONS and JOURNALS}
{Author \headeretal: Preparation of Papers for IEEE TRANSACTIONS and JOURNALS}

\corresp{Corresponding authors: Ashwag Alasmari (aasmry@kku.edu.sa) and Hassan Alhuzali (hrhuzali@uqu.edu.sa)}

\begin{abstract}
Mental health disorders significantly impact people globally, regardless of background, education, or socioeconomic status. However, access to adequate care remains a challenge, particularly for underserved communities with limited resources. Text mining tools offer immense potential to support mental healthcare by assisting professionals in diagnosing and treating patients. This study addresses the scarcity of Arabic mental health resources for developing such tools. We introduce MentalQA, a novel Arabic dataset featuring conversational-style question-and-answer (QA) interactions. To ensure data quality, we conducted a rigorous annotation process using a well-defined schema with quality control measures. Data was collected from a question-answering medical platform. The annotation schema for mental health questions and corresponding answers draws upon existing classification schemes with some modifications. Question types encompass six distinct categories: diagnosis, treatment, anatomy \& physiology, epidemiology, healthy lifestyle, and provider choice. Answer strategies include information provision, direct guidance, and emotional support. Three experienced annotators collaboratively annotated the data to ensure consistency. Our findings demonstrate high inter-annotator agreement, with Fleiss' Kappa of $0.61$ for question types and $0.98$ for answer strategies.  In-depth analysis revealed insightful patterns, including variations in question preferences across age groups and a strong correlation between question types and answer strategies. MentalQA offers a valuable foundation for developing Arabic text mining tools capable of supporting mental health professionals and individuals seeking information.
\end{abstract}


\titlepgskip=-21pt

\maketitle

\section{Introduction}
\label{sec:introduction}

\PARstart{M}{ental} health disorders are highly prevalent worldwide. The impact of mental health can affect individuals regardless of their age, gender, socioeconomic status, or cultural background. According to the World Health Organization, most people do not have access to effective care although one in every eight people in the world experience a mental disorder~\cite{noauthor_undated-ko}. This gap results from several factors, including: a shortage of resources, a low number of mental care professionals, inefficient tools and practices in decision-making, social and cultural taboos~\cite{Zolezzi2018-ld, Kilbourne2018-gk,  Hoffmann2023-fc, Petersen2019-bn}. 

Effective communication is the first step towards building a meaningful relationship between doctor and patient. Yet, Language barriers might become significant obstacles on the path to effective diagnosis and care.  Multiple studies have reported language as a barrier in various health facilities, which have emerged as an alternative to traditional in-person healthcare and services. Especially, during and after the COVID-19 pandemic, health care has been used for frequent consultation as it saves time, resources, and service consumption~\cite{Al_Shamsi2020-su}. The language barrier exists in both traditional and modern clinical settings which impedes effective delivery of medical services. 

The rapid advancement of Artificial Intelligence (AI) and Natural Language Processing (NLP) has also transformed the landscape of disease detection and treatment, including mental illnesses. Innovative tools are being developed to assist mental health care professionals by efficiently reviewing medical history, identifying different patterns, and recommending treatments~\cite{Tutun2023-pf,Van_Heerden2023-sb}.  There has been a significant surge of interest in the development of text mining tools aimed at providing mental health support~\cite{zhang2023phq,Glaz2021-wl,alhuzali2021predicting,Graham2019-wy}. These tools are intended to assist professionals in diagnosing a greater number of patients rather than replacing them. Likewise, by adopting these tools in mental health care, the burden on the health care system can be reduced. However, the development of such tools faces certain limitations, predominantly caused by the scarcity of available datasets.  This challenge is particularly pronounced for the Arabic language, let alone in mental health. 

Existing datasets on mental health are mostly focused on specific disorders, such as suicidal attempts, self-injury, loneliness, depression, or anxiety. This may limit the ability of AI models to diagnose mental health problems.  Examples of those mental health datasets include~\cite{Coppersmith2015-qt,Shen2017-qd,Turcan2019-pt,Rastogi2022-cs,Garg2023-yv}. More specific datasets are being developed that focus on emotions related to specific mental health issues. For example, the CEASE dataset~\cite{ghosh2020cease} focuses on the emotions of people who have attempted suicide, while the EmoMent dataset~\cite{Atapattu2022-xe} focuses on emotions related to depression and anxiety. Other datasets focus on identifying the level of pain in mental health notes such as~\cite{Chaturvedi2023-tg} or identifying the causal interpretation from mental health notes, such as CAMS dataset by~\cite{Garg2022-er}.

Despite efforts worldwide to create corpora in other languages~\cite{Atapattu2022-xe,Kabir2022-jk,Sun2021-nz}, the Arabic language is an understudied language regarding mental health disorders. To date, only a handful of works have considered mental health issues in the Arabic language~\cite{Aldhafer2022-wk,Al-Musallam2022-ut,Al-Laith2021-gy}. In particular, Aldhafer and Yakhlef~\cite{Aldhafer2022-wk} developed depression detection models from Arabic texts on Twitter which focused on the cultural stigma surrounding depression in Arab societies. Another study by Al-Musallam and Al-Abdullatif~\cite{Al-Musallam2022-ut} also focused on the detection of depression in which they applied various machine learning algorithms and feature extraction techniques. 

We observe the following gaps in online mental health research. First, there is a lack of research in Arabic language mental health research and detection. Secondly, most research focused on depression detection in Arabic language, whereas other types of mental health were scarcely researched. The third problem is that most Arabic text research has focused on statements made by people in one-way communication and has ignored the types of questions that arise in two-way communication. 

The objective of our paper is to create a novel Arabic mental health dataset. The obtained corpus comprises a total of 500 questions and answers (Q\&A) posts, including both question types and answer strategies, yielding a total of 1000 annotations. This dataset encompasses interactions, including questions posed by patients and corresponding answers provided by professional doctors. To validate our corpus, we conducted an annotation study following a well-defined annotation schema and employed a quality control process. We also performed extensive analyses to gather evidence on the potential and benefits of our MentalQA dataset. These analyses included correlations between question types and response strategies, an examination of the top frequently used words, an exploration of patient demographics, an analysis of sentiment trends, and an investigation of answering patterns. MentalQA dataset provides valuable resources for constructing effective communication between patients and healthcare providers for mental health support.

The rest of the paper is organized as follows: Section II details the data collection, annotation schema development, and quality control procedures. Section III explores data quality through inter-annotator agreement and analyzes the dataset's characteristics. Section IV discusses the findings and potential applications of MentalQA, acknowledges limitations, proposes future research directions, and explores ethical considerations. Finally, Section V concludes our work.

\section{Methods}
\begin{figure*}
\centering
  \includegraphics[width=0.85\textwidth]{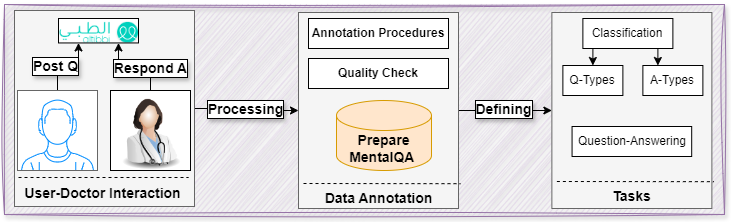}
  \caption{An overview of the creation of Arabic MentalQA dataset, starting from data collection, followed by detailed data annotation and the definition of tasks.}
  \label{framework}
\end{figure*}

This section details the methodology used to construct the MentalQA dataset, a resource for Arabic mental health question answering systems. We collected data from a reputable Arabic medical platform, focusing on mental health questions and answers. An annotation schema was developed to categorize both questions and answers based on their types and strategies. Three expert annotators applied the schema, ensuring consistency through collaboration and independent evaluation. This process resulted in the MentalQA dataset, which offers three key tasks for building future question answering systems in the domain of Arabic mental health. Figure~\ref{framework} presents our framework, illustrating the data format, data annotations, and task definitions that are integral to our approach.

\subsection{Data Collection}

We collected data from the medical platform\footnote{Altibbi.com}, which provides reliable, up-to-date, and simplified medical information in Arabic. The website includes thousands of medical articles, a medical glossary, a Q\&A section, the latest medical news, and additional health services.  For this work, we collected Q\&A posts from 2020 to 2021, as well as the most popular Q\&A posts (i.e., those that were voted as useful by users). This resulted in a dataset of $53,402$ unique Q\&A pairs.  We were interested in mental health questions, so we included only Q\&A posts in the Mental Health category. This process resulted in a final dataset of $2,621$ questions and answers unique pairs. Table 1 presents detailed statistics about the data, including the number of questions, answers, distinct doctors who responded to questions, categories belonging to both questions and answers, and the average length of words per question and answer. 

\begin{table}[ht]
\caption{Data statistics}
\label{tab:stat}
\centering
\begin{tabular}{lc}
\toprule
\textbf{Criteria}                & \textbf{Statistics} \\
\midrule
\textbf{\# Questions (Q)}        & 2,621               \\ 
\textbf{\# Answers (A)}          & 2,621               \\ 
\textbf{\# Categories (Q)} & 7           \\
\textbf{\# Categories (A)} & 3           \\
\textbf{\# Distinct Doctors (A)} & 84           \\
\textbf{\# Avg length words per (Q)} &        30    \\
\textbf{\# Avg length words per (A)} &        31    \\
\bottomrule
\end{tabular}
\end{table}

\subsection{ ANNOTATION SCHEMA DEVELOPMENT}

We based our annotation schema for mental health questions and their corresponding answers on the first layer of the questions classification schema~\cite{Guo2018-px} and answers strategies taxonomy~\cite{Sun2021-nz}, with some modifications. The question classification includes six broad categories, which are described below. It is important to note that the scheme proposed in the study by~\cite{Guo2018-px} consisted of seven general categories. However, the results of our annotation study revealed that one of these categories is not applicable to our dataset. Hence, we excluded it from our annotation schema for that reason.
\begin{itemize}
    \item Diagnosis. Questions about the interpretation of clinical findings, tests, and the criteria and manifestations of diseases.
    
    \item Treatment. Questions about seeking treatments, which may include drug therapy, how to use a drug, and the side effects and contraindications of drugs.

    \item Anatomy and physiology. This category includes important knowledge about basic medicine, such as tissues, organs, and metabolism.

    \item Epidemiology. Questions in this category are mainly about the course, prognosis, and sequelae of diseases, as well as the etiology and causation of diseases, and the association of risk factors with diseases.

    \item Healthy lifestyle. Questions are specified to diet, exercise, mood control and other lifestyle factors that can affect health.

    \item Provider choices. Questions ask for recommendations for hospitals, medical departments, doctors, and the doctor visiting process.

    \item Other. Questions that do not fall under the above-mentioned categories.
\end{itemize}

For answer strategies, some strategies were merged into other categories that can be considered interlinked. For example, we merged the answer strategies of Restatement and Interpretation into the Information category, as they are both likely to seek or provide more information. We also renamed the strategy of Approval and Reassurance to ``Emotional Support'' to make it more inclusive of other types of non-informational support. The Self-disclosure strategy was not applicable due to the nature of our data, as the answers were provided by doctors only. The answers were not evaluated for their completeness or quality with respect to the patient's information needs. The consolidated answer strategies are as follows:

\begin{itemize}
    \item Information. This category includes answers that provide information, resources, etc. It also includes requests for information.

    \item Direct Guidance. This category includes answers that provide suggestions, instructions, or advice. It also includes answers that tell the questioner what they should do to change.

    \item Emotional Support. This category includes answers that provide approval, reassurance, or other forms of emotional support.
\end{itemize}

\subsection{ANNOTATION PROCEDURES}

The annotation process was performed by three annotators who had experience working with biomedical text and natural language processing. They are also native speakers of Arabic. The annotation process started with small batches of 20 questions. As the annotators became more familiar with the task, the batch size was gradually increased to 100 questions. The first 20 questions (trial batch) were the same for all annotators, so they could work on the task in parallel. Their annotations were first checked for quality on the trial batch, and annotators were given feedback to help correct them. Once the annotators had demonstrated that they could produce high quality annotations, they were allowed to work on the main annotation rounds.

The annotation process for the first 200 questions involved a collaborative approach, where all annotators worked simultaneously on the same set of questions. This initial phase aimed to ensure consistency in annotation practices. Before each new batch of data was assigned, the annotators held group meetings to discuss any disagreements that had arisen and to document the resolutions that were agreed upon. Following this collaborative effort, we evaluated the level of agreement among the annotators. The results of this assessment are presented in the Data Quality Check section below. For the remaining 300 questions, we assign each annotator 100 questions to annotate independently. 

\begin{figure*}
\centering
  \includegraphics[width=0.8\textwidth]{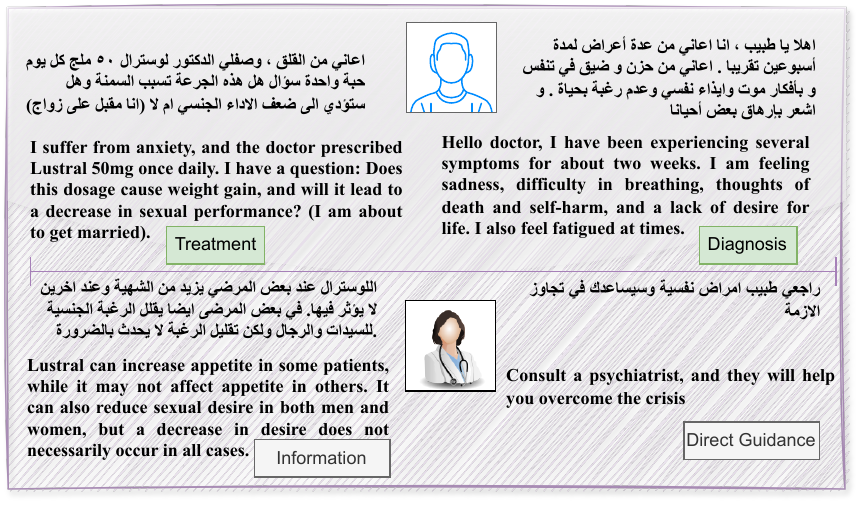}
  \caption{Example of two annotated Q\&A posts, with each Q\&A post translated into English for better readability. The first row represents the questions, while the second row represents the corresponding answers. Additionally, the categories for each question and answer are included.}
  \label{tab:annotated-examples}
\end{figure*}

\subsection{MentalQA Tasks}

The newly created dataset, MentalQA, comprises three interconnected tasks essential for question answering and information retrieval in the context of mental health. The first task involves classifying patients' questions into specific types, facilitating a better understanding of their intent. By categorizing questions based on diagnosis, treatment, anatomy, or others, the dataset enables the development of intelligent question-answering systems tailored to patients' needs. Figure~\ref{tab:annotated-examples} presents two examples with their annotation of both questions and answer types.

The second task focuses on classifying answers into specific strategies, ensuring the extraction of relevant information from a pool of responses. Answer strategies such as informational, direct guidance, or emotional support enhance accuracy and relevance, filtering out unreliable information and providing contextually appropriate answers to patients.

The third task involves developing a question-answering conversational-like style, leveraging the classifications from the previous tasks. This style serves as the backbone for a robust question-answering system, enabling specialized responses to a wide range of mental health-related questions.

In summary, the MentalQA dataset encompasses three tasks: question classification, answer classification, and question-answering conversational-like style. Categorizing questions and answers improves the accuracy, relevance, and reliability of question-answering systems, advancing the field of Arabic mental health and facilitating the development of effective information retrieval systems. MentalQA represents a significant step forward in this domain, paving the way for improved question-answering and information retrieval in mental health settings. In future works, we plan to conduct experiments on these tasks to further validate the effectiveness and applicability of the MentalQA dataset. These experiments will contribute to advancing the field of Arabic mental health and refining question-answering and information retrieval systems for improved patient support and care.

\section{Results}
This section outlines the key findings obtained from analyzing the MentalQA dataset. We assessed inter-annotator agreement, explored the distribution of question types and answer strategies, and investigated the relationship between these elements. Additionally, we analyzed user behavior patterns through gender and age demographics, and explored sentiment within questions and answers. Finally, we examined answering behavior regarding response time, word count, and the language used across different answer strategies. 

\subsection{DATA QUALITY CHECK}

In all the rounds of annotation mentioned, inter-annotator agreement was computed using the Fleiss’ Kappa, a statistical measure for assessing the reliability of agreement between multiple annotators when assigning categorical ratings to a number of items~\cite{fleiss1971measuring}. Table 2 shows the results of our annotation study. The agreement for our annotated Q\&A posts is found to be 0.61 and 0.96 for question types and answers strategies, respectively. We also looked at the total number of questions and answers where all annotators agreed on at least one category, as well as the number of questions and answers where two or more annotators agreed on at least one category. The results showed that (96\%) and (100\%) of the questions and answers, respectively, had at least one category agreed upon by two or more annotators. These findings indicate a substantial level of agreement among annotators in selecting the same category. To generate the final annotations, we employ a majority voting mechanism. 

\begin{table}[ht]
\caption{The findings from our annotation study.}
 \label{tab:annotators}
\centering
\begin{tabular}{p{6cm}c} 
\toprule
\textbf{Criteria} & \textbf{Stat.}  \\ 
\midrule
\# Annotated (Q) & 500 \\ 
\# Annotated (A) & 500 \\  
\# posts where all three annotators annotate	& 200 \\
\# posts where >= 2 annotators agreed on at least 1 category (Q) & 192 \\ 
\# posts where all 3 annotators agreed on at least 1 category (Q) & 100 \\  
\# posts where >= 2 annotators agreed on at least 1 category (A) & 200 \\ 
\# posts where all 3 annotators agreed on at least 1 category (A) & 152 \\ 
\ Fleiss' K (Q)        & 0.61 \\ 
\ Fleiss' K (A)        & 0.96 \\ 
\bottomrule
\end{tabular}
\end{table}

The majority of questions (82\%) had only one label, followed by (18\%) of questions with more than one label. For answers, (69\%) had one label compared to (31\%) which included more than one label. Table~\ref{det_annot} shows the number of questions and answers in each category of question types and answer strategies. According to Table~\ref{det_annot}, treatment is the most common question type (57\%) in the corpus, followed by diagnosis (55\%). Then, health lifestyle and epidemiology question types represent (24\%) and (22\%) of the annotated corpus, respectively. The least represented category in the question types is provider choices (4\%). For answers, the most common answer strategy is information (75\%), followed by direct guidance (56\%) and emotional support (11\%). 

\begin{table}[ht]
\caption{Number of Q\&A posts in each category, where at least two annotators agree on a particular category.}
\label{det_annot}
\centering
\begin{tabular}{lll}
\toprule
\multicolumn{1}{l}{}                    & Categories                   & Counts \\
\midrule
\multirow{6}{*}{Q-types}    & Diagnosis   (A)              & 286    \\
                                          & Treatment   (B)              & 296    \\
                                          & Anatomy   and physiology (C) & 32     \\
                                          & Epidemiology   (D)           & 116    \\
                                          & Healthy   lifestyle (E)      & 125    \\
                                          & Provider   choices (F)       & 20     \\ \midrule
\multirow{3}{*}{A-strategies} & Information   (1)            & 388    \\
                                          & Direct   guidance (2)        & 290    \\
                                          & Emotional   support (3)      & 61    \\
                                          \bottomrule
\end{tabular}
\end{table}

\subsection{ANALYSIS OF QUESTION TYPES AND ANSWERS STRATEGIES}

\begin{figure}[h]
\centering
  \includegraphics[width=\linewidth]{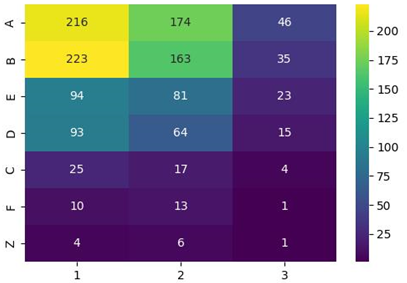}
  \caption{Relationship between Q types and A strategies.}
  \label{fig2}
\end{figure}

Figure~\ref{fig2} reveals intriguing insights into the relationship between the question types and answer strategies. We observe varying degrees of correlation between the categories. For instance, question types A and B demonstrate a strong positive correlation with answer strategies 1 and 2, indicating that these question types are commonly associated with those specific answer strategies. Additionally, question types C, D, and E exhibit moderate positive correlations with answer strategies 1 and 2, indicating a tendency for these question types to be addressed using those strategies. The correlation between answer strategy 3 and question types also demonstrate a noteworthy association. Findings suggest that doctors often leverage emotional support when addressing question type of diagnosis, potentially indicating a tendency to approach the diagnosis process with a compassionate and empathetic approach. By focusing on emotional aspects, doctors may aim to connect with patients. 

\subsection{ANALYSIS OF QUESTION TYPES AND GENDER}
\begin{figure}[h]
\centering
  \includegraphics[width=\linewidth]{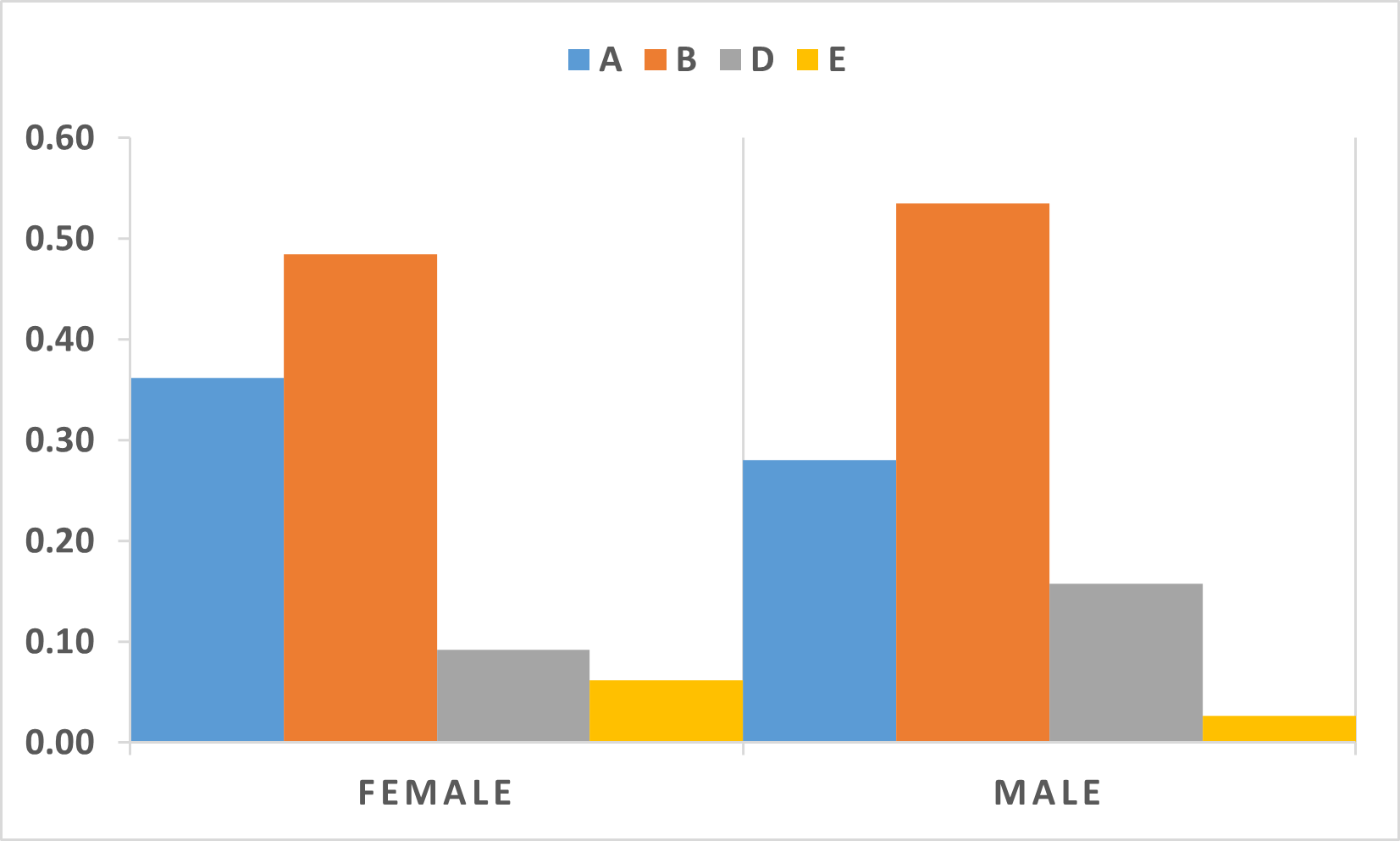}
  \caption{The most asked question types by patients’ gender.}
  \label{fig3}
\end{figure}

Our analysis of the most asked questions by patients' gender, as shown in Figure~\ref{fig3}, has revealed intriguing patterns in the topics they inquire about. This analysis demonstrates that male patients display a higher frequency of questions pertaining to treatment, with a value of 0.54, indicating a strong inclination towards seeking information and guidance on treatment options. In addition, male patients also exhibit an interest in questions about diagnosis, albeit to a lesser extent, with a value of 0.28, suggesting their desire to understand and clarify medical conditions. Conversely, female patients demonstrate a slightly different pattern in their inquiries. Their most prevalent questions predominantly revolve around treatment, with a value of 0.48, indicating a notable emphasis on exploring various treatment methods and approaches. Furthermore, female patients also exhibit a significant interest in questions about diagnosis, with a value of 0.36, highlighting their proactive approach in seeking information regarding medical evaluations. 

\subsection{ANALYSIS OF QUESTION TYPES AND AGE GROUPS}
\begin{figure}[h]
\centering
  \includegraphics[width=\linewidth]{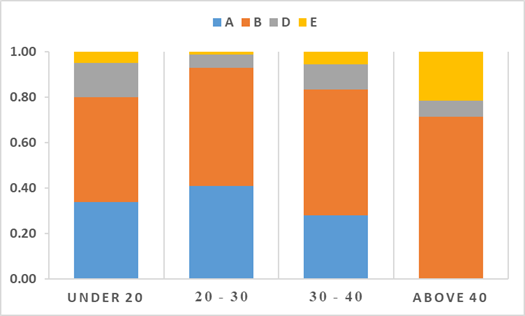}
  \caption{The most asked question types by patients’ age.}
  \label{fig4}
\end{figure}

Our analysis of the most asked question types by patients, as illustrated in Figure~\ref{fig4}, provides valuable insights into the preferences and inquiries across different age groups. The data reveals a breakdown of question types, represented by categories A, B, D, and E, corresponding to specific age groups. Examining the information presented, we observe distinct patterns among the age groups. Patients under 20 years old exhibit a relatively higher proportion of questions in categories A and B, with values of 0.34 and 0.46, respectively. This suggests a high interest in inquiries related to those question types. Conversely, questions falling under categories D and E show lower values for this age group, indicating a relatively lesser focus on those areas. 

Moving to the 20-30 age group, we observe a similar trend. Categories A and B continue to have higher values, with 0.41 and 0.52, respectively, suggesting a sustained interest in those question types. However, there is a notable decrease in the values for categories D and E, indicating a reduced emphasis on those particular question types. For the 30-40 age group, the values for categories A and B decrease further, with 0.28 and 0.56, respectively. This suggests a shift in question preferences compared to the younger age groups. Interestingly, there is a slight increase in the values for categories D and E, indicating a relatively higher interest in those areas within this age group. Finally, the above 40 age group demonstrates a distinct pattern. Category A shows a value of 0.00, indicating a lack of questions related to that specific type. However, categories B, D, and E exhibit higher values, with 0.71, 0.07, and 0.21, respectively. This suggests a significant focus on questions falling within those areas, particularly category B and E.

\subsection{Sentiment Analysis}
\begin{figure}[h]
\centering
  \includegraphics[width=\linewidth]{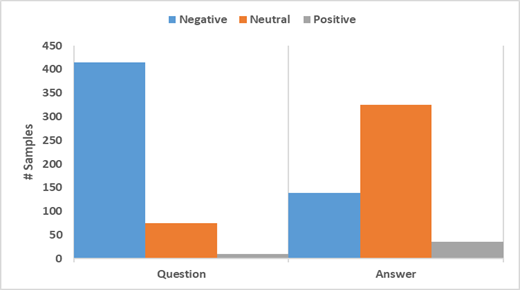}
  \caption{The distribution of sentiment across all annotated posts.}
  \label{fig5}
\end{figure}

We conducted sentiment analysis on the data that yielded intriguing insights as shown in Figure~\ref{fig5}.  Firstly, the analysis revealed that a significant number of patient questions, including 415 posts, were classified as having a negative sentiment. In contrast, 75 posts were labeled as neutral, and only 10 posts were classified as positive. These findings suggest that patients often express negative emotions when discussing their mental health, which could indicate underlying concerns or challenges they face. Secondly, when examining the responses provided by doctors, the sentiment analysis identified a predominant trend of neutrality. Out of the analyzed answers, 325 were labeled as neutral, while 139 were classified as negative, and 36 as positive. This indicates that doctors tend to maintain a neutral tone when addressing patient inquiries, which aligns with their professional approach.  However, it is worth noting that some responses do carry negative emotions, particularly when doctors’ express concerns or fear regarding the patient's health, prompting them to visit a doctor as soon as possible. It should be mentioned that the results of sentiment were computed based on the ``CAMelBERT model developed by Inoue et al.~\cite{Inoue2021-wb}. 

\subsection{ANALYSIS OF Word Frequency}
We have conducted an analysis of word frequency in posts, revealing interesting insights into the associations between different question types and the most frequently used words. Table~\ref{wf} presents the top-15 words that are closely associated with specific question types. This analysis provides valuable information about the common themes and interests expressed by patients when asking different types of questions. For instance, questions related to treatment frequently include words such as “medicine”, “treatment” and “doctor”. Conversely, questions concerning diagnosis often feature words like "feeling", "suffering", and “symptoms”.

\begin{table*}[h]
\centering
\caption{The top 15 words associated with each question’s type. }
\label{wf}

\begin{tabular}{lccccc}
\toprule
\# & A             & B                      & C               & D          & E             \\ \midrule
1  & I feel        & I suffer               & Bloating        & I suffer   & Myself        \\
2  & I suffer      & Medication             & Severe          & Severe     & I feel        \\
3  & When          & Sleep                  & Daily           & Anxiety    & Thoughts      \\
4  & Myself        & I feel                 & Irritable bowel & I feel     & Way           \\
5  & I am          & Myself                 & And the desire  & Thinking   & I know        \\
6  & Sleep         & Treatment              & To defecate     & Disorder   & And feeling   \\
7  & People        & Depression             & More            & Speech     & When          \\
8  & Symptoms      & Doctor                 & 4 times         & Home       & I sleep       \\
9  & Depression    & Treatment              & Per day         & Things     & I have become \\
10 & Heart         & I have                 & Especially      & Personal   & Marriage      \\
11 & I have        & Physician              & Work            & Eating     & Always        \\
12 & Condition     & Fear                   & And I suffer    & Sleep      & Knowledge     \\
13 & Problem       & Doctor                 & Sitting         & I suffer   & People        \\
14 & And lack of   & Best                   & Chair           & Stress     & Situation     \\
15 & Concentration & Problem                & My stomach      & Compulsive & Regret        \\
\bottomrule   
\end{tabular}
\end{table*}

\subsection{ANALYSIS OF ANSWERING BEHAVIOR} 
We also investigated the answering behavior in terms of response time and word frequency.  Table~\ref{tbl5} shows the average answer time and average word counts in answers across different levels of question types in the MentalQA corpus. The results reveal a few interesting patterns. Diagnosis questions (A) are answered quickly (2.61 days) with concise responses (22.33 words). Treatment questions (B) take longer (11.73 days) and have more detailed answers (31.37 words). Notably, questions combining diagnosis and treatment (A, B) take a similar amount of time to answer treatment questions (11.06 days) but require significantly more explanation (47.3 words).

\begin{table}[h]
\caption{Average answer time and average word frequencies in answers across different levels of question types.}
\label{tbl5}
\begin{tabular}{p{2.5cm}cc}
\toprule
Question types & Avg-answer-time & Avg-word-counts\\
\midrule
Diagnosis   (A) & 2.61 & 22.33 \\
Treatment   (B) & 11.73 & 31.37 \\
Diagnosis \& Treatment (A, B) & 11.06 & 47.30 \\
Diagnosis \& Epidemiology (A, D) & 0.30 & 49.07 \\
Diagnosis \& Healthy Lifestyle (A, E) & 11.00 & 27.3 \\
Treatment \& Epidemiology (B, D) & 2.30 & 35.61 \\
Diagnosis, Treatment \& Healthy Lifestyle (A, B, E) & 2.20 & 20.90 \\
\midrule
\end{tabular}
\end{table}

\begin{figure*}[h]
\centering
  \includegraphics[width=0.9\textwidth]{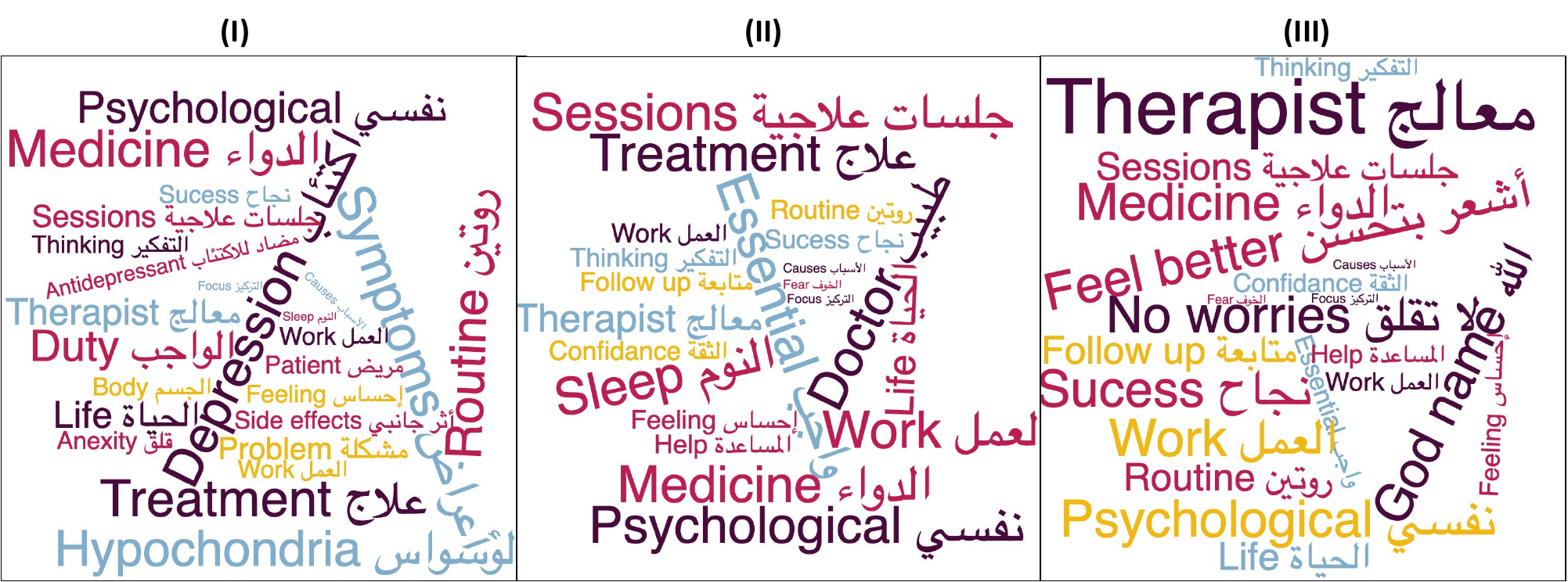}
  \caption{Frequently mentioned words in the answers according to different level of answer strategies where is (I) answers providing only information strategy, (II) answers providing only direct guidance strategy, and (III) answers providing information and emotional support strategy}
  \label{figcloud}
\end{figure*}

To gain deep insights into the answering behavior in MentalQA corpus, we also employed word clouds, as illustrated in Figure~\ref{figcloud}, to visualize the most frequently used words using various answer strategies.  Overall, the word clouds reveal distinct patterns in the terminology used across the different answer strategies. Answers providing solely information frequently employ terms such as ``treatment'', ``depression'', and ``symptoms''. Conversely, answers providing direct guidance predominantly feature terms like ``doctor'', ``essential'', and ``treatment''. Interestingly, answers offering both information and emotional support exhibit a combination of these terms, along with additional words such as ``God name'', ``no worries'', and ``feel better''. 	

\section{DISCUSSION}
In this study, our primary aim was to develop a comprehensive question-answering mental health dataset in Arabic by utilizing posts gathered from an online platform dedicated to mental health. This dataset holds significant value due to the scarcity of research in the field of Arabic language mental health as discussed in~\cite{Zhang2022-la}. To facilitate effective data analysis, we devised an annotation scheme that categorized the posts into six distinct question types and three answer strategies. We conducted an annotation study to validate the reliability of our dataset, which revealed substantial agreement among the annotators. This outcome not only confirmed the high quality of the data but also demonstrated the suitability and applicability of the developed annotation scheme within the context of mental health.

Additionally, we performed extensive analyses to gather evidence on the potential and benefits of our MentalQA dataset. The Results section of this paper presents the findings from various analyses conducted. These analyses included correlations between question types and answer strategies, an examination of the top-18 words used, an exploration of patient demographics, an analysis of sentiment trends, and an investigation of answering behavior. These analyses provide valuable insights into patients' concerns when discussing their mental health issues and how healthcare professionals typically respond to their inquiries.

Furthermore, we conducted an analysis of patient demographics to gain insights into their concerns and priorities, providing valuable information on their specific needs and interests. The higher frequency of treatment-related inquiries from gender and age groups suggests a shared interest in understanding and exploring treatment options. However, variations in the emphasis on diagnosis indicate nuanced differences in their information-seeking behaviors. Understanding these trends enables healthcare providers to tailor their services and communication strategies to better address the specific needs and preferences of different patient groups, ultimately enhancing the quality of care and patient satisfaction. 

Moreover, we conducted analysis on the top-18 words in the dataset. This analysis can help us to capture the collective voice of patients, providing a deeper understanding of the language patterns and preferences that shape their online interactions. For instance, patients used certain words when expressing their mental health concerns on treatment- and diagnosis-related inquiries, which is aligned with the findings of~\cite{Li2023-yg}. Leveraging this knowledge can aid in developing monitoring tools that offer appropriate recommendations to patients and customize content to better meet their needs and interests across various question types.
Also, we performed sentiment analysis to understand the emotional dynamics within patient-doctor interactions~\cite{Li2023-yg, Khanpour2018-kl}. The prevalence of negative sentiments expressed by patients emphasizes the need for empathetic and supportive healthcare practices. Likewise, the predominantly neutral responses from doctors highlight their professionalism while acknowledging occasional displays of negative emotions when warranted by the patient's well-being. These findings contribute to a deeper understanding of the emotional aspects involved in healthcare communication and can inform strategies for improving patient experiences and overall care.

Lastly, we conducted an analysis of answering behavior, which yielded valuable insights into the average response time and word count observed in doctors' replies to patient inquiries. For example, treatment-related questions required more time for doctors to respond compared to diagnosis-related questions. Furthermore, questions containing multiple types tended to be longer, as doctors needed to address multiple inquiries conveyed in the patients' posts.

\subsection{IMPLICATIONS}
MentalQA has the potential to revolutionize mental healthcare access and communication for Arabic-speaking communities. The dataset provides a valuable foundation for developing several tools and resources. For instance, MentalQA can be used to train Large Language Models (LLMs) and other AI models to understand user intent and deliver appropriate responses in Arabic. This paves the way for chat-bots or virtual assistants capable of offering initial mental health support, answering basic questions, and directing users to resources. In addition, Insights from MentalQA can inform the development of communication training materials for healthcare professionals. Additionally, the analysis of question types and corresponding answer strategies can help providers tailor their communication to address patients' specific needs.

MentalQA offers a valuable training resource for LLMs in the domain of Arabic mental health communication. Training LLMs on MentalQA can enhance their ability to understand and respond to Arabic-language mental health inquiries. This can lead to the development of more robust and culturally sensitive AI solutions for mental healthcare applications. Furthermore, MentalQA opens doors for exploring the use of conversational AI models in mental health support services. This could offer a unique way to engage with individuals experiencing mental health challenges and provide them with readily accessible resources.

\subsection{LIMITATIONS}
We used textual data sourced from a consumer health platform to provide a valuable resource for identifying question types and answer strategies within a question-answering platform. However, it is important to note that the user-generated nature of the data introduces limitations to the generalizability of our findings. For instance, not all individuals experiencing mental health issues may have access to the internet or may be unable to express their concerns through such platforms. Consequently, researchers and practitioners utilizing our work should exercise caution in interpreting and applying the results.

For this study, we conducted an annotation study on a total of 500 Q\&A posts, including both question types and answer strategies, resulting in a total of 1000 annotations. The limited availability of resources prevented us from annotating the entire dataset. However, as part of our future work, we plan to annotate the remaining Q\&A posts.

\subsection{ETHICAL CONSIDERATIONS}
To ensure the ethical integrity of our research, we implemented rigorous protocols for data collection from the online platform. Our primary focus was to safeguard the privacy and security of personal data, adhering to strict ethical guidelines. We took extensive measures to anonymize the data, removing any personally identifiable information, and implemented robust data security measures. We are confident that our research will not have any negative ethical implications. However, it is important to note that our analyses may provide valuable insights into the nature of patient inquiries and doctors' responses, allowing for a better understanding of healthcare interactions and potentially leading to improvements in patient care and support.

\section{CONCLUSION}

We have created a novel Arabic mental health dataset. This dataset comprises interactions, which include questions posed by patients and corresponding answers provided by professional doctors. This two-way communication aspect adds immense value to understanding how to effectively support individuals affected by mental health disorders. To ensure the quality and reliability of the dataset, we conducted an annotation study following a well-defined annotation schema and used a quality control process. We also discussed the results of the annotation study and presented statistics outlining the distribution of categories within our dataset. We further included extensive analyses that demonstrate the potential and benefits of our data. In the future, we aim to expand the annotation study to cover the entirety of the data, including the annotation of both questions and answers. Our future goal is to leverage the created dataset to design powerful text mining tools that can make a significant impact in the field of mental health.

\section*{Acknowledgment}
We acknowledge the use of ChatGPT, an AI chatbot developed by OpenAI, for supplementing our own writing, and not to replace them. We verified the accuracy and relevance of the AI-generated text before incorporating it into our manuscript.

\bibliographystyle{IEEEtran}
\bibliography{main}

\begin{IEEEbiography}[{\includegraphics[width=1in,height=1.25in,clip,keepaspectratio]{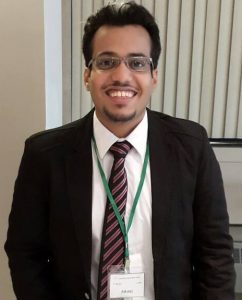}}]{HASSAN ALHUZALI} obtained his Ph.D. degree in Computer Science from the University of Manchester, UK in 2022. Prior to that, he served as an Associate Researcher at the University of Manchester. In 2016, he earned an M.S. degree in Information Science from Indiana University-Bloomington, USA. Following the completion of his master’s degree, he embarked on a period as a visiting student at the Positive Psychology Center at UPENN, USA, as well as UBC, CA. Currently, he serves as an Assistant Professor at Umm Al-Qura University in SA. His ongoing research focuses on natural language processing, affective computing, and mental health.
\end{IEEEbiography}
\begin{IEEEbiography}[{\includegraphics[width=1in,height=1.25in,clip,keepaspectratio]{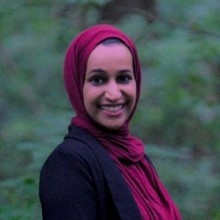}}]{ASHWAG ALASMARI} obtained her PhD degree from University of Maryland, Baltimore County, United States in 2021. She is currently an Assistant Professor in the Department of Computer Science at King Khalid University, Abha, Saudi Arabia. Her research interests include medical question answering, health consumer informatics, human information interaction, natural language processing, and deep learning. Prior to her current position, Dr. Alasmari held research positions at Johns Hopkins University School of Medicine and the U.S. National Library of Medicine (NLM), National Institutes of Health (NIH).
\end{IEEEbiography}
\begin{IEEEbiography}
[{\includegraphics[width=1in,height=1.25in,clip,keepaspectratio]{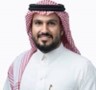}}]{HAMAD ALSALEH} is a faculty member at King Saud University, working within the College of Computer and Information Science, Department of Information Systems. Dr. Alsaleh’s expertise encompasses two key areas: Natural Language Processing (NLP) and Human-Computer Interaction (HCI). With a focus on NLP, Dr. Alsaleh’s explores language understanding and generation, while their work in HCI delves into improving user interaction with technology. Particularly passionate about addressing socio-cultural problems.
\end{IEEEbiography}

\EOD

\end{document}